\documentclass[letterpaper]{article} 
\usepackage{aaai23}  
\usepackage{times}  
\usepackage{helvet}  
\usepackage{courier}  
\usepackage[hyphens]{url}  
\usepackage{graphicx} 
\usepackage{natbib}  
\usepackage{caption} 
\usepackage{algorithm}
\usepackage{algorithmic}

\usepackage{newfloat}
\usepackage{listings}
\DeclareCaptionStyle{ruled}{labelfont=normalfont,labelsep=colon,strut=off} 
\floatstyle{ruled}
\newfloat{listing}{tb}{lst}{}
\floatname{listing}{Listing}
\usepackage{xspace}
\usepackage[nolist,nohyperlinks]{acronym}
\begin{acronym}
\acro{ML}{{Machine Learning}}
\acro{UQ}{{Uncertainty Quantification}}
\acro{CP}{{Conformal Prediction}}
\acro{AI}{{Artificial Intelligence}}
\acro{OD}{{Object Detection}}
\end{acronym}
\newcommand{\ml}{\ac{ML}\xspace}
\newcommand{\uq}{\ac{UQ}\xspace}
\newcommand{\cp}{\ac{CP}\xspace}
\newcommand{\ai}{\ac{AI}\xspace}
\newcommand{\od}{\ac{OD}\xspace}

\usepackage{amssymb}
\usepackage{amsmath}
\usepackage{booktabs}

\newcommand{\dx}{ \widehat{w}^{k}}
\newcommand{\dy}{ \widehat{h}^{k}}
\newcommand{\f}{ \widehat{f}}
\newcommand{\w}{ \widehat{w}}
\newcommand{\h}{ \widehat{h}}

\usepackage{xcolor,cancel}

\title{Conformal Prediction for Trustworthy Detection of Railway Signals}
\author {
    L\'eo And\'eol,\textsuperscript{\rm 1,2}
    Thomas Fel, \textsuperscript{\rm 2,3}
    Florence de Grancey, \textsuperscript{\rm 4}
    Luca Mossina \textsuperscript{\rm 5}
}
\affiliations {
    \textsuperscript{\rm 1} Institut de Math\'ematiques de Toulouse\\
    \textsuperscript{\rm 2} SNCF, 
    \textsuperscript{\rm 3} Brown University, 
    \textsuperscript{\rm 4} Thal\`es\\
    \textsuperscript{\rm 5} IRT Saint Exup\'{e}ry, Toulouse, France\\
    leo.andeol@math.univ-toulouse.fr, thomas\_fel@brown.edu, florence.de-grancey@irt-saintexupery.com luca.mossina@irt-saintexupery.com 
}

\begin{document}
\maketitle

\begin{abstract}
We present an application of conformal prediction, a form of uncertainty quantification with guarantees, to the detection of railway signals.
State-of-the-art architectures are tested and the most promising one undergoes the process of conformalization,
where 
a correction is applied to the predicted bounding boxes (i.e. 
to their height and width) such that they comply with a predefined probability of success.
%
We work with a novel exploratory dataset of images taken from the perspective of a train operator,
as a first step to build and validate future trustworthy machine learning models for the detection of railway signals.

\end{abstract}

\section{Introduction}
In this paper we focus on building a trustworthy detector of railway light signals via \ml\footnote{We refer to ``safety'' and ``trustworthiness'' intuitively; for specific examples, see \citet{alecu_2022_can}.}.
%
We apply \uq to the problem of \od \citep{zhao_2019_object} via the distribution-free, non-asymptotic and model-agnostic framework of \acf{CP}. 
\cp is computationally lightweight, can be applied to any (black-box) predictor with minimal hypotheses and efforts,  and has rigorous statistical guarantees.

We give a brief overview of \uq for \od and the \cp method we apply to our use case.
After selecting a pre-trained model among the state-of-the-art architectures, 
we give some insights on how \uq techniques can quantify the trustworthiness of \od models, which could be part of a critical \ai system, and their potential role in certifying such technologies for future industrial deployment.

\section{The Industrial Problem}
\ai 
can work as a complementary tool to enhance existing technologies, 
like in the automotive industry \citep{alecu_2022_can}.
We can draw a parallel with the railway sector.
While main lines (e.g. high-speed lines) already have in-cabin signaling and can be automatized \citep{singh_2021_deployment}, 
this is too costly to apply to the whole network.
Consequently, 
on secondary lines, drivers 
can be subject to a larger cognitive load, and therefore strain, from 
signals and 
the environment. 
Assisting drivers 
with \ai-based 
signaling recognition could 
facilitate the operations.
%
%

\begin{figure}[t]
    \centering
    \includegraphics[width=0.90\columnwidth]{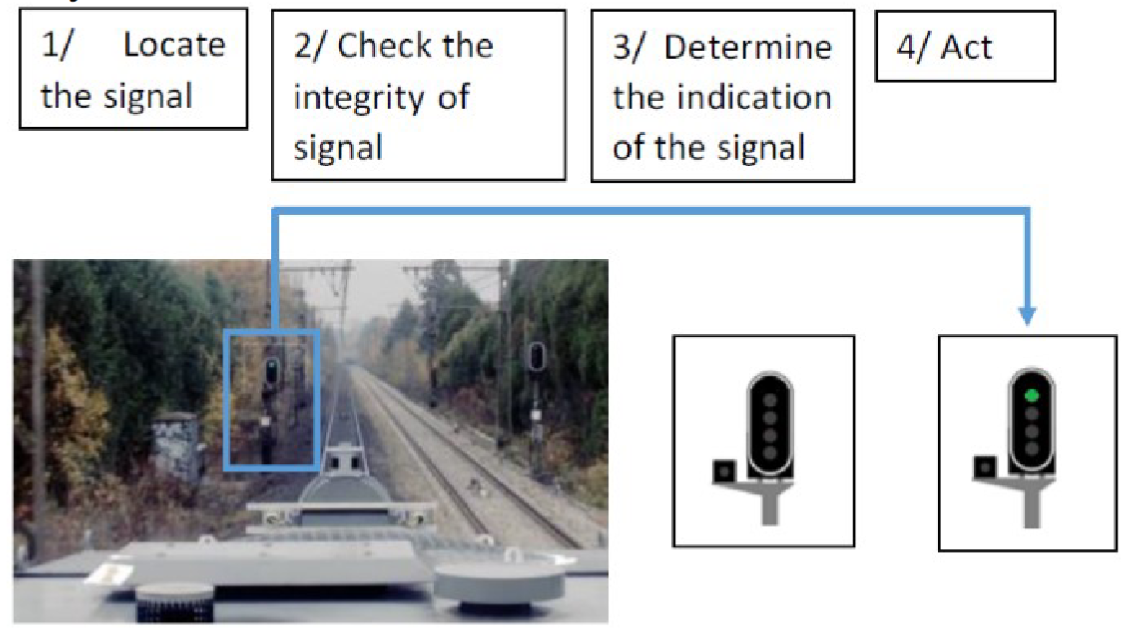}
    \caption{ Example of pipeline where an \ai system acts following \ml-based predictions. Source: \citet{alecu_2022_can}.}
    \label{fig:train-od-pipeline}
\end{figure}

With respect to Figure~\ref{fig:train-od-pipeline}, our application corresponds to point (1) of their process based on multiple \ml tasks: 
trustworthiness concerns can grow with the number of \ml components.
For a recent overview on the technical and regulatory challenges raised by the safety of \ml systems in the railway and automotive industries, see \citet{alecu_2022_can}.

\subsection{Building an exploratory dataset}
Currently, there is no standard benchmark for railway signaling detection.
Stemming from the dataset \textit{FRSign} of \citet{harb_2020_frsign} and their insights, our own iterations made us aware of some additional needs. 
Notably, our new dataset aims for images to be independent and identically distributed (i.i.d), and therefore, raw sequences cannot be considered.
Moreover, our new dataset features increased variability (more railway lines, environmental and weather conditions, etc.) which could enable more accurate predictions in real-world scenarios.
Finally, we generalize the task from single to 
multi-object detection, laying the foundations for future work in instance segmentation. 
We would expect a deployable system to work whenever 
a human operator is successful: at night, with rain, or even when the signals are partially occluded by foliage.
Within our exploratory dataset, we included different light conditions but this is far from exhaustive;
these issues will be taken into account in future 
iterations of the 
dataset. 
%

\begin{table} 
    \centering \small
    \begin{tabular}{l r}
        Characteristics & \multicolumn{1}{c}{Quantity} \\
        \midrule
        Railway lines               & 25 \\
        Images per line (average)   & 55.8 $\pm$ 29.7 \\
        Images in dataset           & 1395  \\
        {Dimensions} (pixels)  & 1280 $\times$ 720  \\
        Bounding boxes (total)      & 2382\\
    \end{tabular}
    \caption{ Characteristics of our dataset} 
    \label{tab:dataset-composition}
\end{table}

As source data, we used some 
video footage taken on French railway lines, with the approval of the creators\footnote{We would like to thank the author of the Youtube channel: \url{https://www.youtube.com/@mika67407}}. 
To extract the samples, we acted as follows.
On 25 videos of average duration about 1.5 hour, we extract on average 55 images per video by running a pretrained object detector with a low objectness threshold, and we keep a minimum interval of 3 seconds between detections, to prevent excessive dependence between images.
We kept the images without the 
associated detections.
We then \textit{manually} annotated all visible railway signals. 
In Table~\ref{tab:dataset-composition} we report the statistics of our dataset.

\section{Related Works}
%
Among the \od architecture, we point out
\textit{YOLO} \citep{redmon2016you} and its variants: 
they propose a one-stage detection combining convolution layers with regression and classification tasks. 
\citet{howard2017mobilenets}, with \textit{MobileNets}, introduced the concept of depth-wise separable convolutions to reduce the number of parameters and accelerate the inference.
As for the recent introduction of transformer layers, 
we find \textit{DETR} \citep{Carion20} and \textit{ViT} \citep{dosovitskiy_2020_image}, among others. 
These networks reach state-of-the-art performances, and seem well-suited to transfer learning. 
%
Finally, \textit{DiffusionDet} \citet{chen2022diffusiondet} recently formulated \od as a denoising diffusion problem, from random noisy boxes to real objects, with state-of-the-art performance.


\subsection{Uncertainty Quantification in Object Detection}
\uq is an important trigger to deploy \od in transport systems.
We find {probabilistic} \od \citep{hall2020probabilistic}, 
where the probability distributions of the bounding boxes and classes are predicted;
{Bayesian models} like in \citet{harakeh2020bayesod} and Bayesian approximations 
\citep{ deepshikha2021monte} are also found in the literature.
%
We point out the distribution-free approach of 
\citet{li_2022_towards}: they build probably approximately correct prediction sets via concentration inequalities, estimated via a held-out calibration set.
They control 
the coordinates of the boxes 
but also the proposal and objectness scores, resulting in 
more and larger boxes. 
%
Finally, \citet{degrancey_2022_detection} proposed an extension of \cp to \od, which will be the framework of choice in our exploratory study.
 
\section{Conformalizing Object Detection}
\acf{CP} \citep{vovk_2022_alrw, angelopoulos_2021_gentle} 
is a family of methods to perform \uq with guarantees under the sole hypothesis of data being independent and identically distributed (or more generally exchangeable). 
%
For a specified 
(small) error rate $\alpha \in (0,1)$, at inference, the \cp procedure will yield prediction sets
$C_{\alpha}(X)$ that contain the true target values $Y$ with probability: 
%
%
%
\begin{equation}
    \mathbb{P} \big( Y_{new} \in C_{\alpha}(X_{new}) \big) \geq 1 - \alpha. \label{eq:cp-guarantee}
\end{equation}

This guarantee holds true, on average, over all images at inference and over many repetitions of the \cp procedure.
However, it is valid for any distribution of the data $\mathbb{P}_{XY}$, any sample size and any predictive model $\f$, even if it is misspecified or a black box.
The probability $1-\alpha$ is the \textit{nominal coverage} and the \textit{empirical coverage} on $n_{\text{test}}$ points is $\sum_{i=1}^{n_{\text{test}}} \mathbb{I}\{Y_i \in {C}_{\alpha}(X_i)\} / n_{\text{test}}$. 
%

We focus on \textit{split} \cp \citep{papadopoulos_2002_inductive, lei_2018_distribution},  
%
where one needs a set of \textit{calibration} data $D_{cal}$ drawn from the same $\mathbb{P}_{XY}$ as the test data, with no need to access the 
training data.
At \textit{conformalization}, 
we compute 
the \textit{nonconformity scores}, to quantify the uncertainty on held-out data. 
At inference, \cp adds a margin around the bounding box predicted by a pretrained detector $\f$.

\subsubsection{Split conformal object detection}
We follow \citet{degrancey_2022_detection}.
Let $k=1, \dots, n_{box}$ index every ground-truth box in $D_{cal}$ that was detected by $\f$,
{disregarding} their source image. 
Let $Y^{k} = (x_{\text{min}}^{k}, y_{\text{min}}^{k}, x_{\text{max}}^{k}, y_{\text{max}}^{k})$ be the 
coordinates of 
the $k$-th box and $\widehat{Y}^{k} = (\hat{x}_{\text{min}}^{k}, \hat{y}_{\text{min}}^{k}, \hat{x}_{\text{max}}^{k}, \hat{y}_{\text{max}}^{k})$ its prediction. 
%
 
In \citet{degrancey_2022_detection} their nonconformity score, which we refer to as \textit{additive}, is defined as: 
\begin{equation} \small
    R_{k} = ( 
     \hat{x}_{\text{min}}^{k} -     {x}_{\text{min}}^{k}, 
     \hat{y}_{\text{min}}^{k} -     {y}_{\text{min}}^{k},
         {x}_{\text{max}}^{k} - \hat{x}_{\text{max}}^{k},
         {y}_{\text{max}}^{k} - \hat{y}_{\text{max}}^{k}).
     \label{eq:score-additive}
\end{equation}

\noindent We further define the \textit{multiplicative} one as:
\begin{equation} \small
    R_{k} = \big(
          \frac{\hat{x}_{\text{min}}^{k} -     {x}_{\text{min}}^{k}}{\dx} ,
          \frac{\hat{y}_{\text{min}}^{k} -     {y}_{\text{min}}^{k}}{\dy} ,
           \frac{    {x}_{\text{max}}^{k} - \hat{x}_{\text{max}}^{k}}{\dx}, 
           \frac{    {y}_{\text{max}}^{k} - \hat{y}_{\text{max}}^{k}}{\dy} \big)
    \label{eq:score-multiplicative}
\end{equation}
where the prediction errors are scaled by the predicted width $\dx$ and height $\dy$. 
This is similar to \citet{papadopoulos_2002_inductive}, and a natural extension to \citet{degrancey_2022_detection}.

\textit{Split conformal object detection} goes as follows:

\begin{itemize} \itemsep0em
    \item[1.] choose a nonconformity score: e.g. Equation~(\ref{eq:score-additive}) or (\ref{eq:score-multiplicative});
    \item[2.] Run a \textit{pairing mechanism}, to associate predicted boxes with true boxes (see following paragraphs);
    \item[3.] For every coordinate $c \in \{x_\text{min},y_\text{min},x_\text{max},y_\text{max}\}$, let $\bar{R}^{c} = \left(R_{k}^{c}\right)_{k=1}^{n_{box}}$; 
    \item[4.] compute $q_{1 - \frac{\alpha}{4}}^{c} = \lceil (n_{box}+1)(1-\frac{\alpha}{4}) \rceil$-th element of the {\it sorted} sequence $\bar{R}^{c}$, $\forall c \in \{x_\text{min},y_\text{min},x_\text{max},y_\text{max}\}$. 
\end{itemize}

Since we work 
with four coordinates, for statistical reasons, we adjust the quantile order from $(1-\alpha)$ to $(1- \frac{\alpha}{4})$ using the Bonferroni correction.
%
%
%
%
We conformalize \textit{box-wise}:
we want to be confident that when we detect correctly an object (``true positive''), we capture the entire ground-truth box with a frequency of at least $(1-\alpha)$, on average.
%
%
During calibration (point 2. above), for all the true positive predicted boxes ($\widehat{Y}^{TP}_{cal}$), 
we compute the nonconformity score between the true box and the prediction.
%
The {pairing mechanism} 
is the same as the NMS used in \od, 
that is, for each ground-truth bounding box, the predicted bounding boxes (not already assigned to a ground truth) are tested in decreasing order of their confidence scores. 
The first predicted box with an IoU above a set threshold is assigned to the ground truth box.
Note that while building ($\widehat{Y}^{TP}_{cal}$), we do not consider false negatives 
(due to $\f$) 
which cannot be taken care of by box-wise \cp.

At inference, the \textit{additive} split conformal object detection prediction box is built as:
\begin{align} \small
    \widehat{C}_{\alpha}(X) = \{ 
        \widehat{x}_{\text{min}} &- q_{1-\frac{\alpha}{4}}^{x_\text{min}},\,
        \widehat{y}_{\text{min}}  - q_{1-\frac{\alpha}{4}}^{y_\text{min}}, \\ 
        \widehat{x}_{\text{max}} & + q_{1-\frac{\alpha}{4}}^{x_\text{max}},\,  
        \widehat{y}_{\text{max}}  + q_{1-\frac{\alpha}{4}}^{y_\text{max}} \}. \nonumber
    \label{eq:add-pred-set}
\end{align}

\noindent The \textit{multiplicative} conformal prediction box is:
\begin{align} \small
    \widehat{C}_{\alpha}(X) = \{ 
        \widehat{x}_{\text{min}} &- \w \cdot q_{1-\frac{\alpha}{4}}^{x_\text{min}},\,
        \widehat{y}_{\text{min}}  - \h \cdot q_{1-\frac{\alpha}{4}}^{y_\text{min}},\\
        \widehat{x}_{\text{max}} &+ \w \cdot q_{1-\frac{\alpha}{4}}^{x_\text{max}},\,
        \widehat{y}_{\text{max}}  + \h \cdot q_{1-\frac{\alpha}{4}}^{y_\text{max}} \}. \nonumber 
    \label{eq:mul-pred-set}
\end{align}

\newcommand{\yolo}{YOLOv5\xspace}
\newcommand{\ddet}{DiffusionDet\xspace}
\section{Experiments} \label{sec:experiments}
{We split our dataset into three subsets: validation, calibration and test (respectively of size 300, 700, 395)}.
We compare, using {the validation set}, the performance of \yolo, DETR and \ddet, pretrained on COCO \citep{lin_2014_coco}, 
as candidate base predictors $\f$, restricting the detection to the class ``traffic light''.
%
Commonly, \od models 
are 
evaluated using the Average Precision (AP), which is the Area Under the recall-precision Curve.
AP 
incorporates the precision-recall trade-off, and the best value is reached when precision and recall are maximized for all objectness thresholds.
%
In our application (Table~\ref{tab:detector-metrics}), the AP for \yolo is low, 
while DETR gives better results, and 
\ddet is 
significantly superior to the others.
We therefore use the DiffusionDet model for our conformalization.
However, AP metrics alone do not give a complete picture to select \od architecture. 

\begin{table} 
    \centering \small
    \begin{tabular}{ l c c}

    &  \multicolumn{2}{c}{Average Precision} \\
        \cmidrule(lr){2-3}
     & IoU $\geq 0.3$ & IoU $\geq 0.8$\\
    \midrule
    YOLOv5s$^{\dag}$          & 0.239 & 0.033  \\
    YOLOv5x$^{\dag}$          & 0.287 & 0.093 \\
    DETR resnet50    & 0.531 & 0.008   \\
    DiffusionDet     & \textbf{0.839} & \textbf{0.325} \\
    \end{tabular}
    \caption{Comparing models via {AP} for two IoU threshold levels. {$^{\dag}$: v5s and v5x respectively correspond to a small and a large configurations of YOLOv5.}}
    \label{tab:detector-metrics} 
\end{table}

\begin{figure}[h!]
    \centering
    \includegraphics[width=0.95\columnwidth]{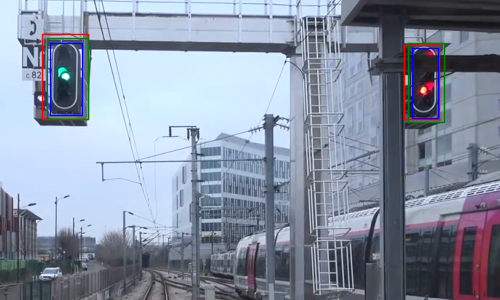}
    \caption{Bounding boxes as predicted by the \ml predictor (blue), their conformalized version with \textit{additive} scores (green) and the ground truth (red). Example of images used in conformalization and testing, cropped for readability.}
    \label{fig:cp-additive}
\end{figure}

\subsubsection{Detector evaluation with box-wise conformalization}
\begin{figure}[h!]
    \centering 
    \includegraphics[width=0.95\columnwidth]{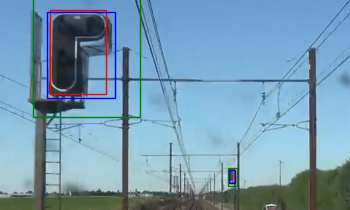}
    \caption{Bounding boxes as predicted by the \ml predictor (blue), their conformalized version with \textit{multiplicative} scores (green) and the ground truth (red). Example of images used in conformalization and testing, cropped for readability. 
    }
    \label{fig:cp-multiplicative}
\end{figure}

We evaluate \od performances (Table~\ref{tab:cp-margins-diffusion}) by box-wise conformal prediction, {reporting} the estimated quantiles $q_{1-\frac{\alpha}{4}}^{c}$ at the desired risk level.
A higher quantile reveals a higher uncertainty of \od predictions.In order to compare additive and multiplicative quantiles which are qualitatively different, we report the {\it stretch}: the ratio of the areas of the conformalized and predicted boxes. We display examples of additive and multiplicative conformalized boxes respectively on Figure \ref{fig:cp-additive} and Figure \ref{fig:cp-multiplicative}.
\begin{table} 
    \centering \small
    \begin{tabular}{ l cccc c}
      & $q_{1-\alpha}^{x_\text{min}}$ & $q_{1-\alpha}^{y_\text{min}}$ & $q_{1-\alpha}^{x_\text{max}}$ & $q_{1-\alpha}^{y_\text{max}}$  & Stretch$^{*}$ \\ 
    \midrule
    Additive $^\dag$         & 4.74 & 7.16 & 6.91 & 5.67  & $\times$ 2.857 \\
    Multiplicative $^\ddag$  & 0.22 & 0.37 & 0.21 &  0.21 & $\times$ 2.259 \\
    \end{tabular}
    \caption{Conformalization margins on DiffusionDet.
    $^*$: average increase in the {area of the boxes} 
    after conformalization. $^\dag$: pixels; $^\ddag$: fraction of $\w$ or $\h$.}
    \label{tab:cp-margins-diffusion}
\end{table}

%

%


\section{Results} \label{sec:results}
%
Empirically, \cp works as expected: in Table~\ref{tab:cp-coverage}, we see that conformalized coverage is close to the nominal level of $(1 - \alpha) = 0.9$.
Remark that the guarantee holds on average over multiple repetitions, hence we cannot expect to attain the {requested} nominal coverage with just one dataset. 
\begin{table} 
    \centering \small
    \begin{tabular}{l c c c }
    \cp method &   \multicolumn{1}{c}{None}  & Additive & Multiplicative  \\
    \midrule
    Empirical coverage & 0.106 & 0.921 & 0.894 \\
    \end{tabular}
    \caption{Empirical coverage before and after conformalization, nominal coverage: $1 - \alpha = 0.90$. This replies to: ``how many true boxes, when detected, were entirely covered by the predicted box?''}
    \label{tab:cp-coverage}
\end{table}


\subsection{Interpretation of conformal bounding boxes}

A conformalized predictor is only as good as its base predictor $\f$.
If the latter misses many ground truth boxes, guaranteeing $(1 - \alpha) \, 100 \%$ correct predictions of a few boxes will still be a small number.
%
That is, conformalization is not a substitute for careful training and fine-tuning of a detection architecture, but a complementary tool for certification.

The interest of capturing the whole box can be operational: our \ml pipeline could rely on a conservative estimation of the ground-truth to carry out a control operations (e.g. running a \ml subcomponent on the detection area).


\section{Conclusion \& perspectives}

Given the insights from this exploratory study, we plan on building and publishing an augmented version of the dataset.
The objective is to have a dedicated, high-quality benchmark for the scientific community and the transport industry.
As mentioned above, \cp works with exchangeable data.
In the long term, if trustworthy \ai components are to be deployed, the \uq guarantees will need to be adapted to streams of data: this will pose a theoretical challenge and one in the construction and validation of the dataset.

So far, the underlying criterion for successful prediction has been whether the ground truth box is \textit{entirely} covered by the predicted box.
This is strict, as having a system that guarantees to cover a big part of the truth, seems to be equally useful in practice.
\citet{bates_2021_rcps}, with their \textit{risk controlling prediction sets}, and \citet{angelopoulos_2022_conformal_risk}, with \textit{conformal risk control}, go in this direction.
They extend the guarantee of \cp to arbitrary losses, that can incorporate other operational needs.

\bibliography{aaai23}

\end{document}